\documentclass[10pt,twocolumn,letterpaper]{article}

\usepackage{cvpr}
\usepackage{times}
\usepackage{epsfig}
\usepackage{graphicx}
\usepackage{amsmath}
\usepackage{amssymb}
\usepackage{mathrsfs}
\usepackage{float} 
\usepackage{CJK} 
\usepackage{amsmath} 
\usepackage{amssymb} 
\usepackage{longtable} 
\usepackage{multirow} 
\usepackage{array} 
\usepackage{chngpage} 
\usepackage{mathrsfs} 
\usepackage{float}

\usepackage{url}
\usepackage{marvosym}
\usepackage{eurosym}
\usepackage{filecontents}
\usepackage{booktabs}

\usepackage{algorithm}
\usepackage{algorithmicx}
\usepackage{algpseudocode}

\usepackage{caption}
\usepackage{subcaption}
\usepackage{cite}

\usepackage[breaklinks=true,bookmarks=false]{hyperref}

\cvprfinalcopy 

\begin{document}

\title{Scene Text Recognition with Sliding Convolutional Character Models}

\author{Fei Yin, Yi-Chao Wu, Xu-Yao Zhang, Cheng-Lin Liu\\
National Laboratory of Pattern Recognition, Institute of Automation, 
Chinese Academy of Sciences\\
95 Zhongguan East Road, Beijing, 100190, PR. China\\
{\tt\small \{fyin, yichao.wu, xyz, liucl\}@i1.org}
}

\maketitle

\begin{abstract}
	Scene text recognition has attracted great interests from the computer vision and pattern recognition community in recent years. State-of-the-art methods use concolutional neural networks (CNNs), recurrent neural networks with long short-term memory (RNN-LSTM) or the combination of them. In this paper, we investigate the intrinsic characteristics of text recognition, and inspired by human cognition mechanisms in reading texts, we propose a scene text recognition method with character models on convolutional feature map. The method simultaneously detects and recognizes characters by sliding the text line image with character models, which are learned end-to-end on text line images labeled with text transcripts. The character classifier outputs on the sliding windows are normalized and decoded with Connectionist Temporal Classification (CTC) based algorithm. Compared to previous methods, our method has a number of appealing properties: (1) It avoids the difficulty of character segmentation which hinders the performance of segmentation-based recognition methods; (2) The model can be trained simply and efficiently because it avoids gradient vanishing/exploding in training RNN-LSTM based models; (3) It bases on character models trained free of lexicon, and can recognize unknown words. (4) The recognition process is highly parallel and enables fast recognition. Our experiments on several challenging English and Chinese benchmarks, including the IIIT-5K, SVT, ICDAR03/13 and TRW15 datasets, demonstrate that the proposed method yields superior or comparable performance to state-of-the-art methods while the model size is relatively small.
\end{abstract}

\section{Introduction}

With the development of the Internet and widespread use of mobile devices with digit cameras, there are massive images in the world and many of them contain texts. The text in natural image carries high level semantics and can provide valuable cues about the content of the image. Thus, if texts in these images can be detected and recognized by computers, they can play significant roles for various vision-based applications, such as spam detection, products search, recommendation, intelligent transportation, robot navigation and target geo-location. Consequently, scene text detection and recognition has become a hot research topic in computer vision and pattern recognition in recent years.

\begin{figure}
	\centering
	
	\includegraphics[height=7cm,width=7.5cm]{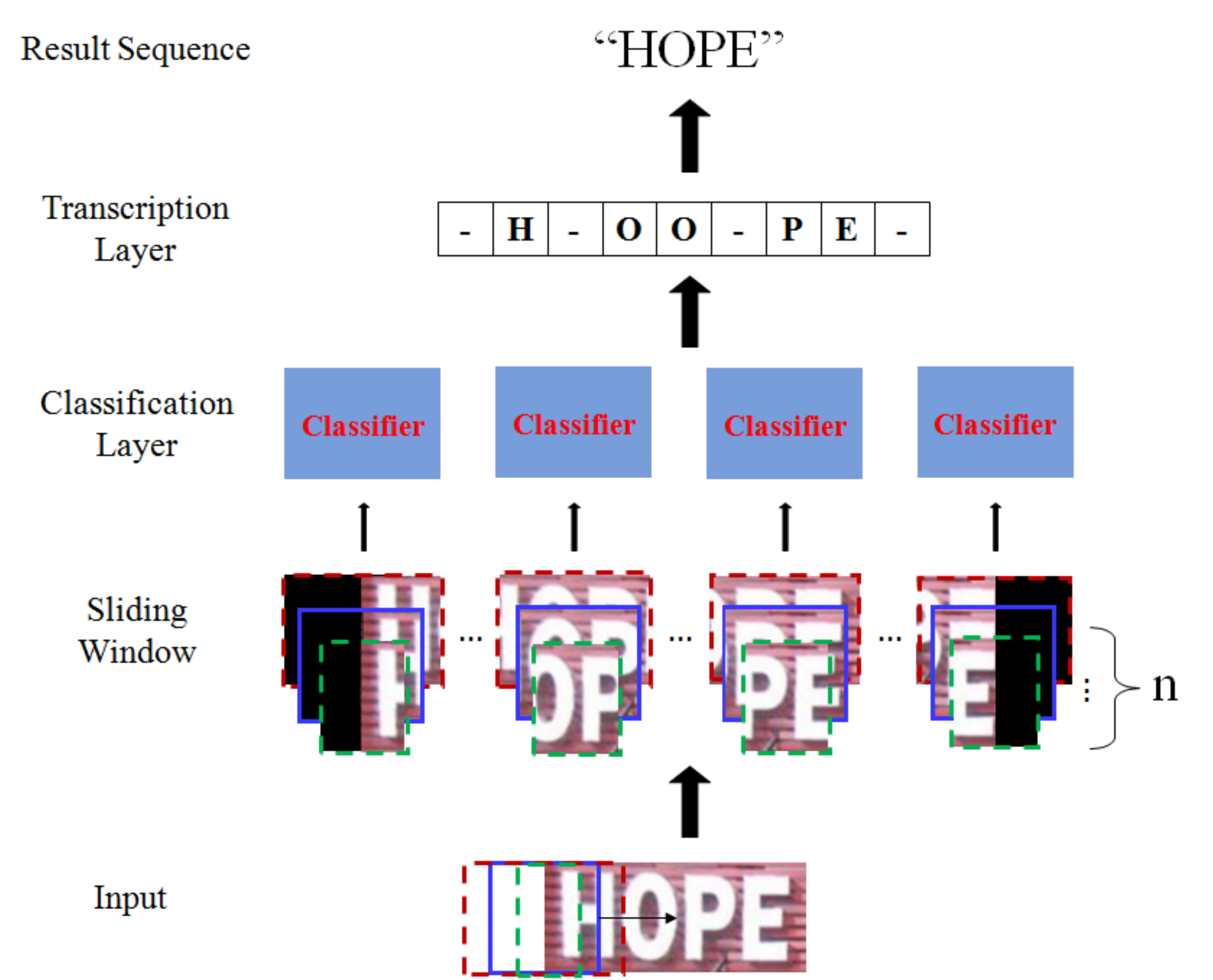}
	
	\caption{The framework of the proposed method. The framework consists of three parts: 1) Sliding window layer, which extract features from the window; 2) Classification layer, which predicts a label distribution from the input window image; 3) Transcription layer, which translates the per-window predictions into the final label sequence.}\label{fig:FrameWork}
\end{figure}

Although traditional Optical Character Recognition (OCR) has been investigated for a few decades and great advances have been made for scanned document images \cite{1, 2}, the detection and recognition of text in both natural scene and born-digital images, so called robust reading, remains an open problem \cite{3, 4}. Unlike the texts in scanned document images which are well-formatted and captured under a well-controlled environment, texts in scene images are largely variable in appearance and layout, drawn from various color, font and style, suffering from uneven illumination, occlusions, orientations, distortion, noise, low resolution and complex backgrounds (Fig. 2). Therefore, scene text recognition remains a big challenge.

\begin{figure}
	\centering
	
	\includegraphics[height=2cm,width=8.1cm]{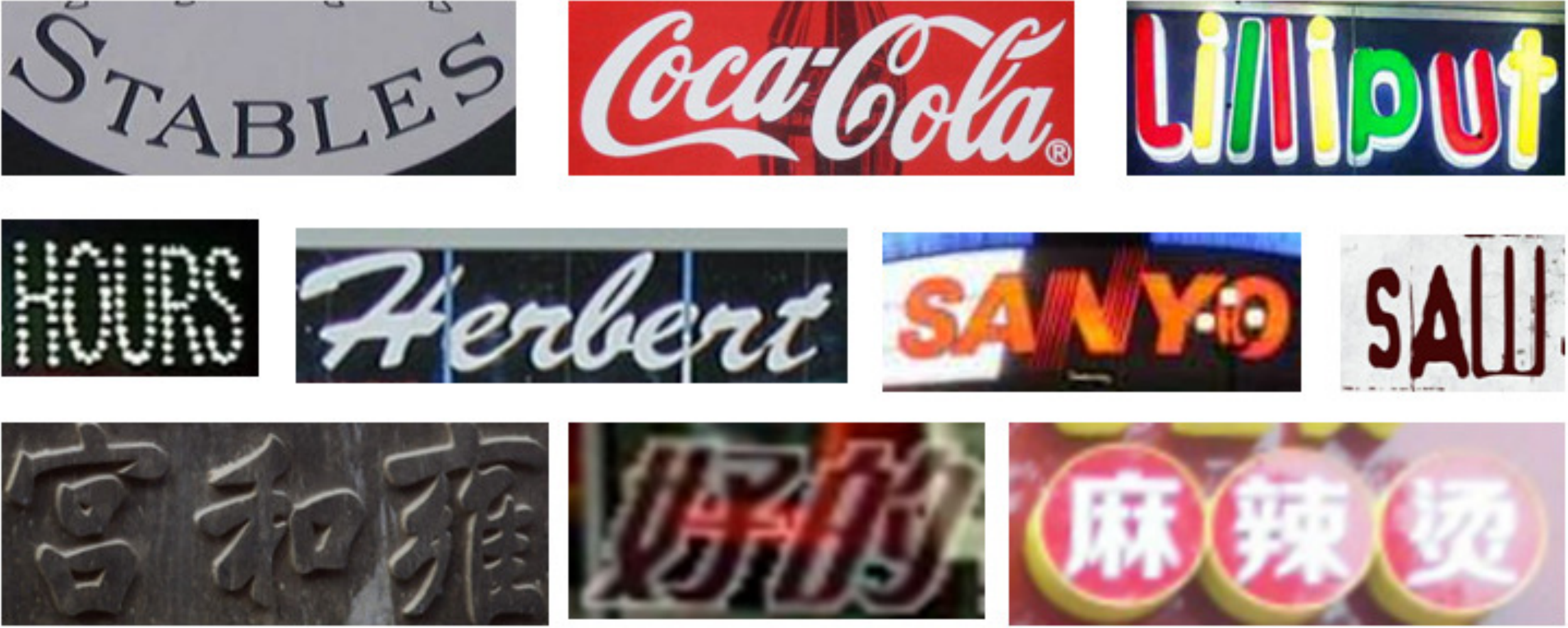}
	
	\caption{Examples of English and Chinese Scene Text}\label{fig:TE5KD}
\end{figure}

Many efforts have been devoted to the difficult problem of scene text recognition. The methods so far can be roughly categorized into three groups: explicit segmentation methods,  implicit segmentation methods and holistic methods. (1) Explicit segmentation methods\cite{5,6,7,8,9,10,11,12,13,14} usually involve two steps: character segmentation and word recognition. It attempts to segment the input text image at the character boundaries to generate a sequence of primitive segments with each segments being a character or part of a character, applies a character classifier to candidate characters and combine contextual information to get the recognition result. Although this approach has performed well in handwritten text recognition, the performance in scene text recognition is severely confined by the difficulty of character segmentation. However, explicit segmentation methods have good interpretation since they can locate the position and label of each character. (2) Implicit segmentation methods \cite{15,16,17,18,19,20,21} regard text recognition as a sequence labeling, which avoids the difficult character segmentation problem by simply slicing the text image into frames of equal length and labeling the sliced frames. Hidden Markov Model (HMM) and Recurrent Neural networks (RNNs) are typical examples for this case. In particular, the combination of convolutional neural network (CNN) and RNN based network obtained the state-of-the-art results on several challenging benchmarks. However, RNN-based methods have two demerits: (a) The training burden is heavy when the input sequence is very long or the number of output classes is large; (b) The training process is tricky due to the gradient vanishing/exploding etc. (3) Holistic methods \cite{22,23,24,25,26,27,28,29} recognize words or text lines as a whole without character modeling. Though this is feasible for English word recognition and has reported superior performance, its reliance on a pre-defined lexicon makes it unable to recognize a novel word, And also, holistic methods are not applicable to the case that fixed lexicon is not possible, e.g., for Chinese text line recognition.

Despite the big progress in recent years, the current scene text recognition methods are insufficient in both accuracy and interpretation compared to human reading.  Modern cognitive psychology research points out that reading consists of a series of saccades (whereby the eyes jump from one location to another and during which the vision is suppressed so that no new information is acquired) and fixations (during which the eyes remain relatively stable an process the information in the perceptual span) \cite{30}, Therefore, if we simplify the perceptual span  as a window, the process of reading can be formulated by a sliding window which outputs the meaningful recognition results only when its center is at the fixation point.

Based on the above, we propose a simple and efficient scene text recognition method inspired by human cognition mechanisms, in which a sliding window and a character classifier based on deep neural network are used to imitate the mechanisms of saccades and fixations, respectively. Our method has several distinctive advantages: 1) It simultaneously detects and recognizes characters and can be trained on weakly labeled data;  2) It achieves competitive performance on both English and Chinese scene text recognition; 3) The recognition process is highly parallel and enables fast recognition. We evaluate our method on a number of challenging scene text datasets. Experimental results show that our method yields superior or comparable performance compared to the state of the art.

The rest of the paper is organized as follow. Section 2 reviews related works. Section 3 details the proposed method. Experimental results are given in Section 4 and conclusion is drawn in Section 5.

\section{Related Work}

In recent years, a large number of scene text recognition systems have been reported in the literature, and some representative methods are reviewed below.  

In general, explicit segmentation methods consists of character segmentation and word recognition. The recognition performance largely relies on character segmentation. The existing segmentation methods roughly fall in two categories: binarization based and detection based. Binarization based methods find segmentation points after binarization. Niblack's adaptive binarization and Extremal Regions (ERs) are two typical binarization based methods, which are employed in \cite{12} and \cite{11}, respectively. However, since text in natural scene image suffers from uneven illumination and complex backgrounds, binarization can hardly give satisfactory results. Detection based methods bypass the binarization by adopting multi-scale sliding window strategy to get candidate characters from the original image directly. For example, the methods in \cite{6,8,9} directly extract features from the original image and use various classifiers to decide whether a character exist in the center of a sliding window. Shi et al. \cite{10} employ a part-based tree-structured model and a sliding window classification to localize the characters in the window. Detection based methods overcome the difficulty of character segmentation and have shown good performance.

In explicit segmentation based methods, the integration of contextual information with character classification scores is important to improve the recognition performance. The methods of integration include Pictorial Structure models (PS) \cite{6}, Bayesian inference \cite{7}, and Conditional Random Field (CRF)\cite{8,9,10}. Wang et al. \cite{6} use the PS to model each single character and the spatial relationship between characters. This algorithm shows good performance on several datasets, but it can only handle words in a pre-defined dictionary. Weinman et al. \cite{7} proposed a probabilistic inference model that integrates similarity, language priors and lexical decision to recognize scene text. Their approach was effective for eliminating unrecoverable recognition errors and improving accuracy. CRF is was employed in \cite{8, 9} to jointly model both bottom-up (character) and top-down (language) cues. Shi et al. \cite{10} built a CRF model on the potential character locations to incorporate the classification scores, spatial constraints, and language priors for word recognition. In \cite{11}, word recognition was performed by estimating the maximum a posterior (MAP) under joint distribution of character appearance and language prior. The MAP inference was performed with Weighted Finite-State Transducers (WFST). Moreover, beam search has been used to achieve fast inference \cite{12, 13} for overcoming the complexity with high-order context models (e.g., 8th-order language model in \cite{12}).

Implicit segmentation methods take the whole image as the input and are naturally free from the difficulty of character segmentation which severely hinders performance of the explicit segmentation methods. With the use of deep neural network, implicit segmentation methods have shown overwhelming superiority in scene text recognition. The implicit segmentation methods use either hand crafted features \cite{17} or features learned by CNN \cite{18, 19, 20, 21}, the labeling algorithm is either HMM \cite{15, 16} or LSTM \cite{17, 18, 19, 20, 21}. Recently, the combination of CNN and LSTM has led to state-of-the-art performance \cite{19}. 

In holistic recognition methods, Goel et al. \cite{22} use whole word sub-image features to recognize the word by comparing to simple black-and-white font-renderings of lexicon words. The methods in \cite{23, 24, 25} use word embedding, in which the recognition becomes a nearest neighbor classification by creating a joint embedding space for word images and the text. In \cite{26, 27}, Jaderberg et al. develop a powerful convolutional neural network to recognize English text by regarding every English word as a class. Thanks to the strong classification ability of CNN and the availability of large set of training images by synthesis, this method shows impressive performance on several benchmarks. Holistic recognition is confined by a pre-defined lexicon, however, although Jaderberg et al. \cite{28} and Lee et al. \cite{29} propose another CNN based model which can recognize unconstrained words by predicting the character at each position in the output text, it is highly sensitive to the non-character space.

The proposed method is an implicit segmentation method. It overcomes the difficulty of character segmentation by sliding window, and the underlying CNN character model can be learned end-to-end with training images weakly labeled with text scripts only.

\section{The Proposed Method}

The  framework of the proposed method, as shown in Fig. 1, consists of three parts: a sliding window layer, a classification layer and a transcription layer.

The sliding window layer extracts features from the window, and the features can be the original image with different scale hand-crafted features or CNN features. On the top of sliding window, a classifier is built to predict a label distribution from the input features. The transcription layer is adopted to translate the per-window predictions into the result sequence. The whole system can be jointly optimized as long as the classifier is differentiable, making the back propagation algorithm workable.

\subsection{Rationale}

When humans read a text line, their eyes do not move continuously along a line of text, but make short rapid movements intermingled with short stops. During the time that the eye is stopped, new information is brought into the processing, but during the movements, the vision is suppressed so that no new information is acquired \cite{30}. Inspired by these, we build our scene text recognition system which follows a simplified process of human reading. We assume it only skips one character in each saccade as a unskilled people, then we use exhaustive scan windows with suitable step to imitate the saccade. When the centre of the scan window coincide with the fixation point, the character classifier outputs character labels and confidence scores, otherwise, it outputs 'blank'.

For the window size, we consider the fact that after height normalization, characters usually have approximately similar width in the text image. Therefore, we fix the size of the sliding window to a right size in which a character can be covered completely. Though fixing the window size may bring disturbance to character classification as shown in Fig3, we can see the the character is still recognizable when it is in the center of window for printed character and Chinese handwritten character(e.g., Fig3 (b), Fig3(c) and Fig3(d)).

\begin{figure}
	\centering
	\begin{subfigure}[b]{0.20\textwidth}
		\centering
		\includegraphics[width=\textwidth,height=0.7\textwidth]{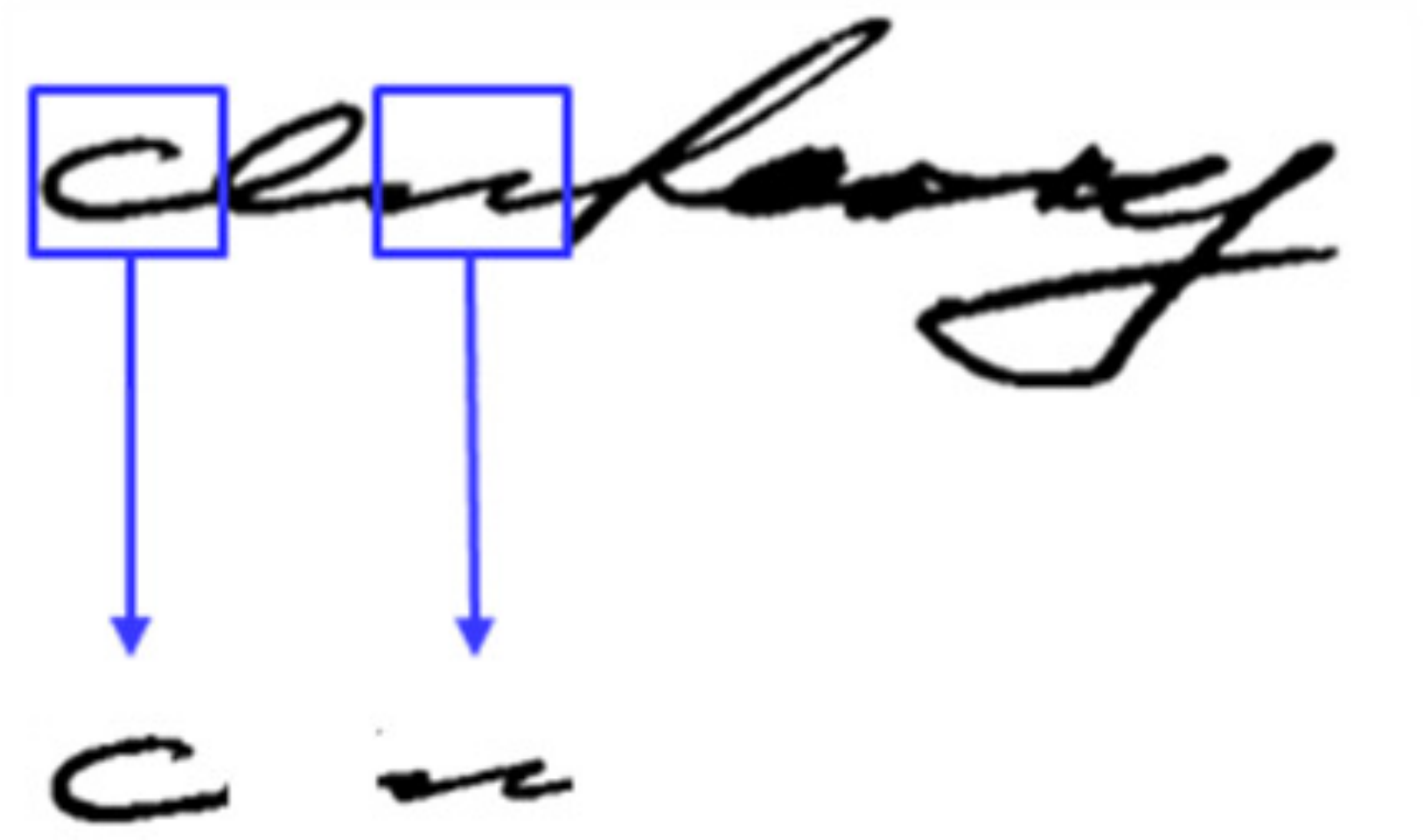}
		\caption{~}
	\end{subfigure}
	\begin{subfigure}[b]{0.20\textwidth}
		\centering
		\includegraphics[width=\textwidth,height=0.7\textwidth]{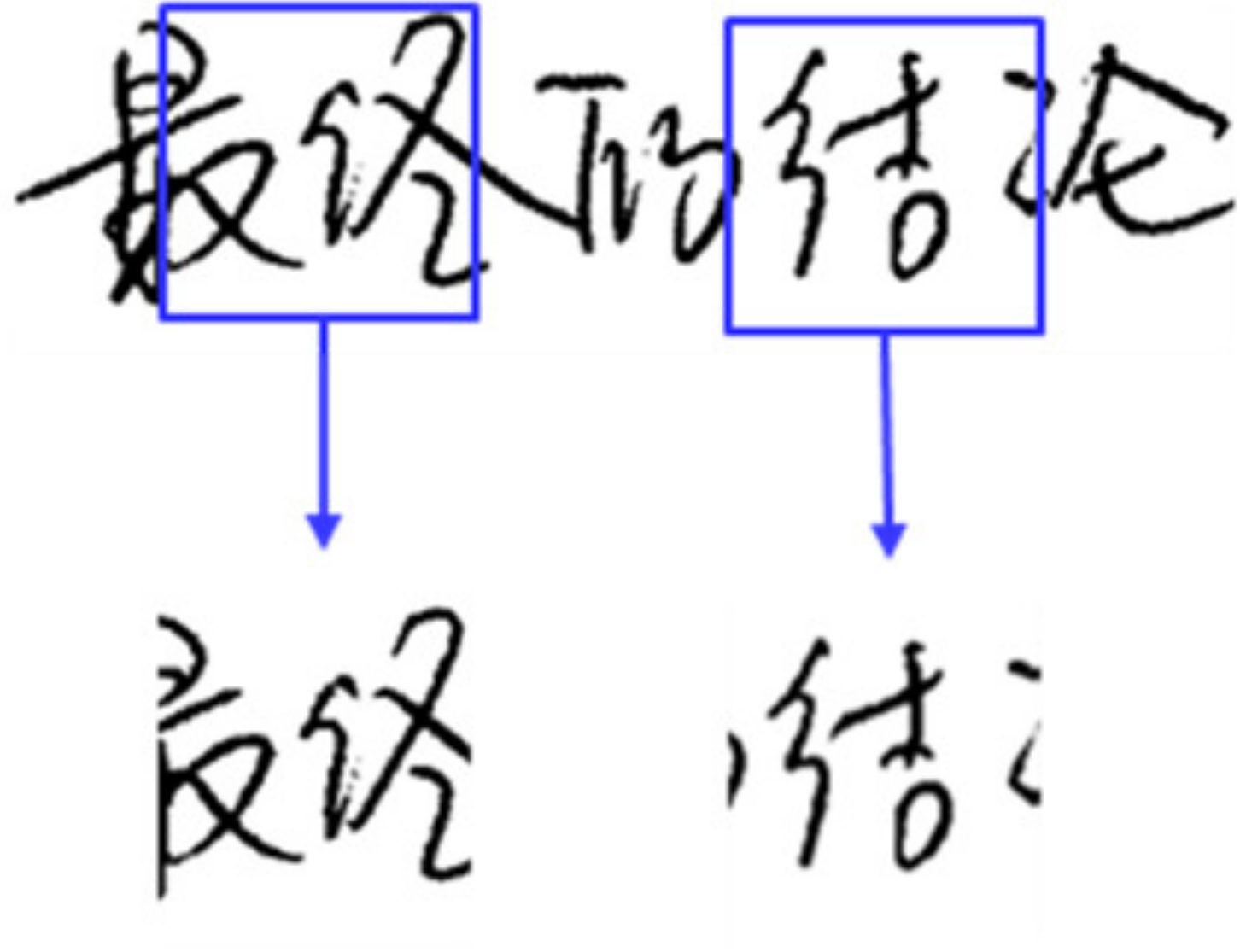}
		\caption{~}
	\end{subfigure}
	\vskip\baselineskip
	\begin{subfigure}[b]{0.20\textwidth}
		\centering
		\includegraphics[width=\textwidth,height=0.7\textwidth]{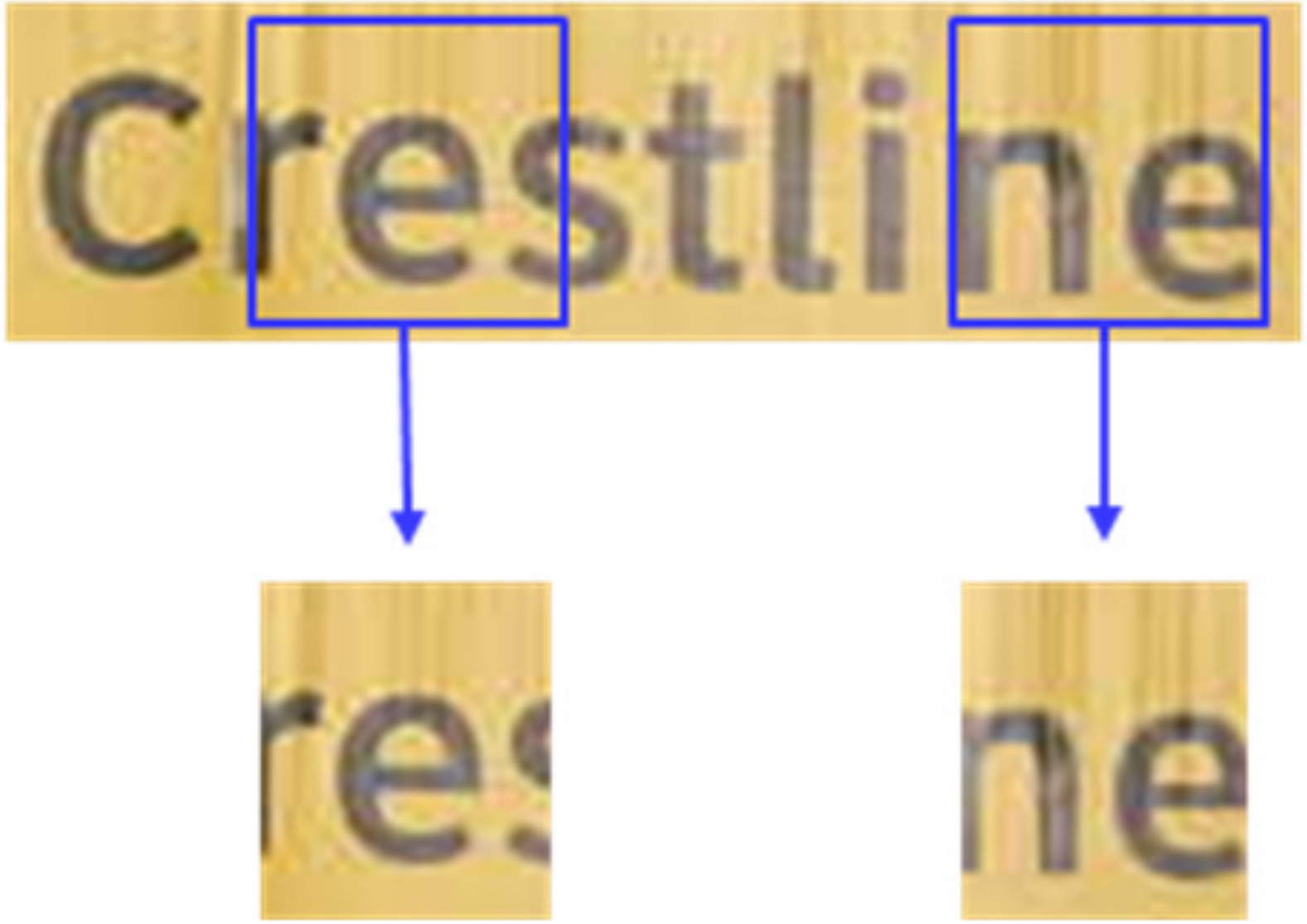}
		\caption{~}
	\end{subfigure}
	\quad
	\begin{subfigure}[b]{0.20\textwidth}
		\centering
		\includegraphics[width=\textwidth,height=0.7\textwidth]{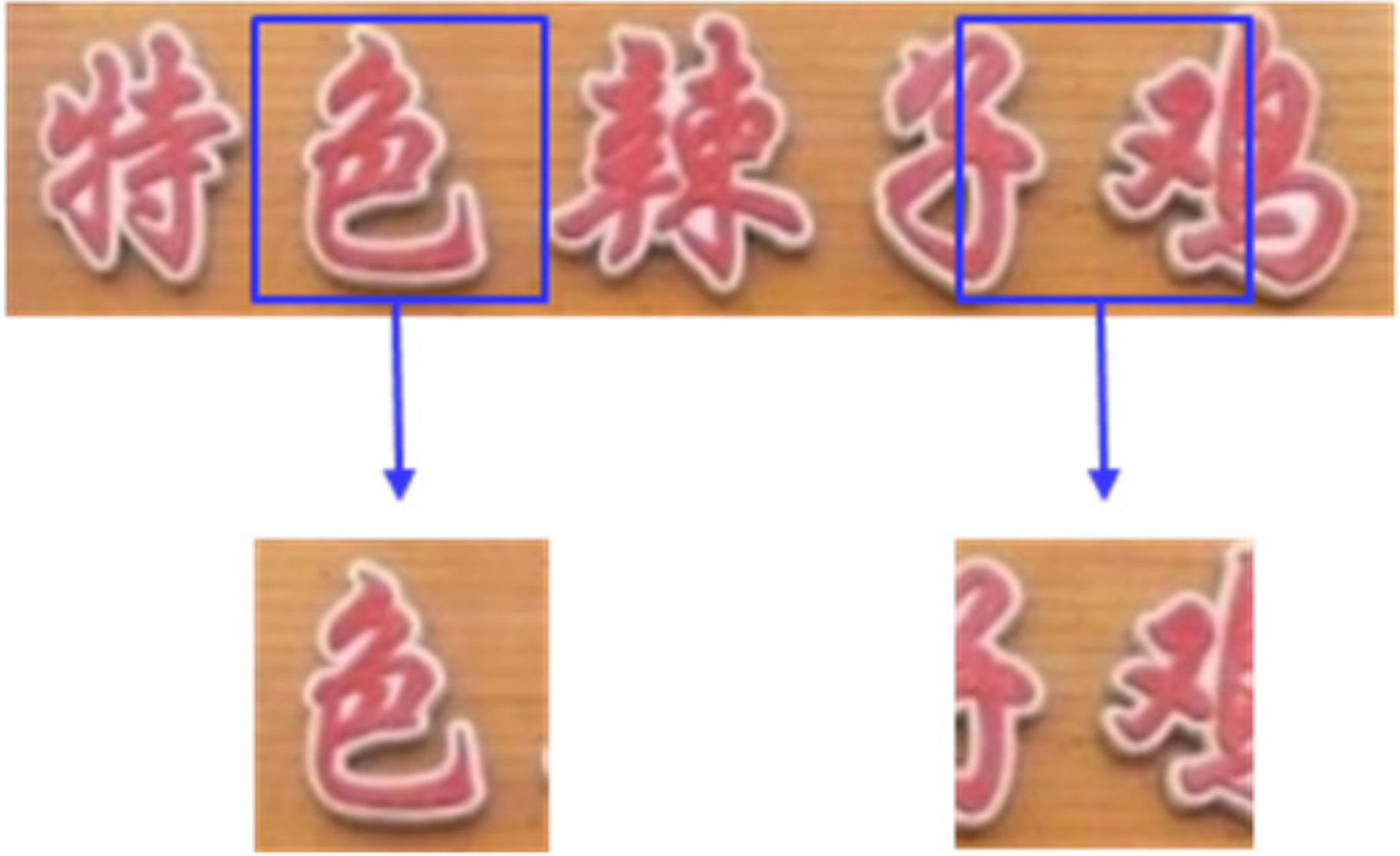}
		\caption{~}
	\end{subfigure}
	\caption[ ]
	{\small Some examples of printed scene text and hadwritten text. (a) Handwritten English; (b) Handwritten Chinese; (c) Printed Scene English Text; (d) Printed Scene Chinese Text.}
	\label{fig:mean and std of nets}
\end{figure}

Our framework is flexible.  If we reduce the number of window to one, extend the size of window to whole image and use the crnn model to recognize the context in the window, our framework degenerates to the model in \cite{19}.  If we reduce the width of window to one pixel and model the relationship between windows, our framework becomes a RNN based model.

Our framework also relates to previous methods in \cite{31, 32}, where neural nets were trained on digit string images. However, our method differs in that it can be trained on the weakly labeled datasets, i.e., without the need of locating characters in training images. As we know,  labeling locations of characters is laborious and time consuming, which makes training with big data infeasible.

\subsection{Convolutional Character Model}

Although our text recognition framework use any classifier for character recognition on sliding window, in this work, we use the CNN which has been proven superior in recent results \cite{33, 34}. We build a 15-layer CNN as the character model as shown in Table 1, which is similar to the one proposed in \cite{34}. We resize the original character image in the sliding window to $32 \times 32$ as the input feature map. The filters of convolutional layers are with a small receptive field $3 \times 3$, and the convolution stride is fixed to one. The number of feature maps is increased from 50 to 400 gradually. To further increase the depth of the network so as to improve the classification capability, spatial pooling is implemented after every three convolutional layers instead of two \cite{34}, which is carried out by max-pooling (over a $2 \times 2$ window with stride 2) to halve the size of feature map. After the stack of 12 convolutional layers and 4 max-pool layers, the feature maps are flattened and concatenated into a vector with dimensionality 1600. Two fully-connected layers (with 900 and 200 hidden units respectively) are then followed. At last, a sofmax layer is used to perform the 37/7357-way classification. To facilitate CNN training, we adopt the batch normalization (BN) technique, and insert eight BN layers after some of the convolution layers.


\begin{table}[!htbp]
	\centering
	\caption{CNN configuration. 'k', 's', 'p' stand for kernel size, stride and padding size, respectively.}
	\scalebox{0.80}[0.80]{
		\begin{tabular}{|c|c|}\hline
			Type & Configuration \\ \hline
			Input & N*32*32 gray-scale images \\ \hline	
			Convolution & \#maps:50, k:3*3, s:1, p:1, drop:0.0 \\ \hline
			Batch Normalization & \\ \hline	
			Convolution & \#maps:100, k:3*3, s:1, p:1, drop:0.1 \\ \hline
			Convolution & \#maps:100, k:3*3, s:1, p:1, drop:0.1 \\ \hline
			Batach Normalization & \\ \hline	
			Max-pooling & Windows:2*2, s:2 \\ \hline
			
			Convolution & \#maps:150, k:3*3, s:1, p:1, drop:0.2 \\ \hline
			Batch Normalization & \\ \hline	
			Convolution & \#maps:200, k:3*3, s:1, p:1, drop:0.2 \\ \hline
			Convolution & \#maps:200, k:3*3, s:1, p:1, drop:0.2 \\ \hline
			Batach Normalization & \\ \hline	
			Max-pooling & Windows:2*2, s:2 \\ \hline
			
			Convolution & \#maps:250, k:3*3, s:1, p:1, drop:0.3 \\ \hline
			Batch Normalization & \\ \hline	
			Convolution & \#maps:300, k:3*3, s:1, p:1, drop:0.3 \\ \hline
			Convolution & \#maps:300, k:3*3, s:1, p:1, drop:0.3 \\ \hline
			Batach Normalization & \\ \hline	
			Max-pooling & Windows:2*2, s:2 \\ \hline
			
			Convolution & \#maps:350, k:3*3, s:1, p:1, drop:0.4 \\ \hline
			Batch Normalization & \\ \hline	
			Convolution & \#maps:400, k:3*3, s:1, p:1, drop:0.4 \\ \hline
			Convolution & \#maps:400, k:3*3, s:1, p:1, drop:0.4 \\ \hline
			Batach Normalization & \\ \hline	
			Max-pooling & Windows:2*2, s:2 \\ \hline
			
			Fully Connection & \#hidden units:900, drop:0.5 \\ \hline
			Fully Connection & \#hidden units:200, drop:0.0 \\ \hline
			
			Softmax & \#output units:37/7357 \\ \hline
			
		\end{tabular}
	}
	
\end{table}

\subsection{Transcription}

Transcription is to convert the per-window predictions made by the convolutional character model into a sequence of character labels. In this work, we assume each window represents a time step, and then adopt the CTC layer as our transcription layer.

CTC \cite{1} maximizes the likelihood of an output sequence by efficiently summing over all possible input-output sequence alignments, and allows the classifier to be trained without any prior alignment between input and target sequences. It uses a softmax output layer to define a separate output distribution \(P(k|t)\) at every step \(t\) along the input sequence for extended alphabet, including all the transcription labels plus an extra ‘blank’ symbol which represents an invalid output. A CTC path \(\pi \) is a length \(T\) sequence of blank and label indices. The probability \(P(\pi |{\rm{X}})\)  is the emission probabilities at every time-step:

\begin{equation}
P(\pi |{\rm{X}}) = \sum\limits_{t = 1}^T {P({\pi _t}|t,{\rm{X}})}.
\end{equation}

Since there are many possible ways of separating the labels with blanks, to map from these paths to the transcription, a CTC mapping function \(B\) is defined to firstly remove repeated labels and then delete the ‘blank’ from each output sequence. The conditional probability of an output transcription \(y\)  can be calculated by summing the probabilities of all the paths mapped onto it by \(B\):

\begin{equation}
P({\rm{y}}|{\rm{X}}) = \sum\limits_{\pi  \in {B^{ - 1}}({\rm{y}})} {P{\rm{(}}\pi {\rm{|X)}}}.
\end{equation}

To avoid direct computation of the above equation, which is computationally expensive, we adopt, the forward-backward algorithm \cite{1} to sum over all possible alignments and determine the conditional probability of the target sequence.

\subsubsection{Decoding}

Decoding a CTC network means to find the most probable output transcription \({\rm{y}}\) for a given input sequence \({\rm{X}}\). In practice, there are mainly three decoding techniques, namely naive decoding, lexicon based decoding and language model based decoding. In naive decoding, predictions are made without any lexicon or language model.
While in lexicon based decoding, predictions are made by search within a lexicon. As for the language model based method, a language model is employed to integrate the linguistic information during decoding.


\textbf{Naive Decoding}: The naive decoding, which also refer to as best path decoding without any lexicon or language model, is based on the assumption that the most probable path corresponds to the most probable transcription:

\begin{equation}
\begin{array}{l}
{y^*} \approx B({\pi ^*}), \\ 
{\pi ^*} = \arg {\max _\pi }P(\pi |{\rm{X}}), \\ 
\end{array}
\end{equation}

\textsc{N}aive decoding is trivial to compute, since \({\pi ^*}\)  is just the concatenation of the most active outputs at every time-step.


\textbf{Lexicon Based Decoding}: In lexicon-based decoding, we adopt the technique called token passing proposed in \cite{1}. First, we add ‘blank’ at the beginning and end and between each pair of labels. Then a token \(tok(s,h)\) is defined, where \(s\) is a real-valued score and \(h\)  represents previously visited words. Therefore, each token corresponds to a particular path through the network outputs, and the token score is the log probability of that path. At every time step \(t\)  of the output sequence with length \(T\), each character \(c\)  of word \(w\)  holds a token \(tok(w,c,t)\), which is the highest scoring token reaching that segment at that time. Finally, the result can be acquired by the \(tok(w, - 1,T)\) . The details can be found in \cite{1}. In our implementation, to make the recognition result contains only one word, we set the score of the token with more than one history words to be extremely small.


\textbf{Language Model Based Decoding}: Statistical language model only models the probabilistic dependency between adjacent characters in words, which is a weaker linguistic constraint than lexicon. In language model based decoding, we adopt the refined CTC beam search by integration of language model to decode from scratch, which is similar to the one proposed in \cite{35} . We denote the blank, non-blank and total probabilities assigned to partial output transcription  \(y\)  of time \(t\)  as  \({\Pr ^ - }({\rm{y}},t)\), \({\Pr ^ + }({\rm{y}},t)\) and \(\Pr ({\rm{y}},t)\), respectively. The extension probability \(\Pr (k,{\rm{y}},t)\) of  \(y\) by label \(k\)   at time \(t\)  is defined as follows:

\begin{equation}
\Pr (k,{\rm{y}},t) = \Pr (k,t|{\rm{X}}){P^\alpha }(k|{\rm{y}})\left\{ {\begin{array}{*{20}{c}}
	{{P^ - }({\rm{y}},t - 1){\rm{ ~if~ }}{{\rm{y}}^{\rm{e}}} = k}  \\
	{P({\rm{y}},t - 1){\rm{ ~otherwise}}}  \\
	\end{array}} \right.
\end{equation}

\noindent where \(\Pr (k,t|{\rm{X}})\) is the CTC emission probability of \(k\)  at \(t\) ,  \(y^e\) is the final label of \(y\), \({P}(k|{\rm{y}})\) is the linguistic transition from \(y\)  to \({\rm{y}} + k\) and can be re-weighted with parameter \(\alpha \). The search procedure is described in Algorithm 1. Our decoding algorithm is different from that of \cite{35} in two aspects. First, we introduce a hyper-parameter \(\alpha \) to the expression of extension probability, which accounts for language model weight. Second, in order to further reduce the search space, we prune emission probabilities at time \(t\)  and retain only the top candidate number (\(CN\)) classes.

\begin{algorithm}
	\caption{Refined CTC Beam Search}
	\begin{algorithmic}[1] 
		\State \textbf{Initialize:} $B \gets \{\phi\}$; $Pr^{-}(\phi,0)=1$
		\For{$i = 1 \to T$}
		\State $\hat{B} \gets the~N$-$best~sequences~in~B$
		\State $B \gets \{\}$
		\For{$y \in \hat{B}$}
		\If {$y \neq \phi$}
		\State $Pr^{+}(y,t) \gets Pr^{+}(y,t-1)Pr(y^{e},t|X)$
		\If {$\hat{y} \in \hat{B}$}
		\State $Pr^{+}(y,t) \gets Pr^{+}(y,t)+Pr(y^{e},y,t)$
		\EndIf
		\EndIf
		\State $Pr^{-}(y,t) \gets Pr^{-}(y,t-1)+Pr(-,t|X)$
		\State add~$y$~to~$B$
		\State sort~emission~probabilities~at~time~$t$~and~retain~top~$CN$~classes
		\For {$k = 1 \to CN$}
		\State $Pr^{-}(y+k,t) \gets 0$
		\State $Pr^{+}(y+k,t) \gets Pr(k,y,t)$
		\State add~$y$~to~$B$
		\EndFor
		\EndFor
		\EndFor
		\State \textbf{Return:}$\max_{y \in B}Pr^{\frac{1}{\|y\|}}(y,T)$
	\end{algorithmic}
\end{algorithm}

\subsection{Model Training}
Denote the training dataset by $D=\{X_i, Y_i\}$, where $X_i$ is a training image of word or text line and $Y_i$ is the ground truth label sequence. The objective is to minimize the negative log-likelihood of conditional probability of ground truth:

\begin{equation}
O =  - \sum\limits_{{X_i},{Y_i} \in D} {\log p({Y_i}|{S_i})}.
\end{equation}

\noindent where $S_i$ is the window sequence produced by sliding on the image $X_i$. This objective function calculates
a cost value directly from an image and its ground truth label sequence. Therefore, the network can be end-to-end
trained on pairs of images and sequences, eliminating the procedure of manually labeling all individual characters
in training images. The network is trained with stochastic gradient descent (SGD) implemented by Torch 7\cite{36}.

\begin{table*}[!htb]
	\centering
	\caption{Recognition accuracies (\%) on four English scene text datasets. In the second row, “50”, “1k” and “Full” denote the lexicon used, LM denotes the language model and “None” denotes recognition without  language constraints. (\textbf{*\cite{28} is not lexicon-free in the strict sense, as its outputs are constrained to a 90k dictionary.})}
	\scalebox{0.80}[0.75]{
		\begin{tabular}{cccccccccccccccccc}\toprule
			~ & \multicolumn{4}{c}{\textbf{IIIT5k}} & ~ & \multicolumn{3}{c}{\textbf{SVT}} & ~ & \multicolumn{4}{c}{\textbf{IC03}} & ~ & \multicolumn{2}{c}{\textbf{IC13}} \\
			\cmidrule(lr){2-5}
			\cmidrule(lr){7-9}
			\cmidrule(lr){11-14}
			\cmidrule(lr){16-17}
			
			\textbf{Method} & \textbf{50} & \textbf{1k} & \textbf{LM} & \textbf{None} & ~ & \textbf{50} & \textbf{LM} & \textbf{None} & ~ & \textbf{50} & \textbf{Full} & \textbf{LM} & \textbf{None} & ~ &  \textbf{LM} & \textbf{None} & \textbf{Model Size} \\ \midrule
			ABBYY\cite{6}  & 24.3 & - & - & - & ~ & 35.0 & - & - & ~ & 56.0 & 55.0 & - & - & ~ & - & - & - \\
			Wang et al.\cite{6} & - & - & - & - & ~ & 57.0 & - & - & ~ & 76.0 & 62.0 & - & - & ~ & - & - & - \\
			Mishra et al.\cite{9} & 64.1 & 57.5 & - & - & ~ & 73.2 & - & - & ~ & 81.8 & 67.8 & - & - & ~ & - & - & - \\
			Novikova et al.\cite{11} & - & - & - & - & ~ & 72.9 & - & - & ~ & 82.8 & - & - & - & ~ & - & - & - \\
			Wang et al.\cite{14} & - & - & - & - & ~ & 70.0 & - & - & ~ & 90.0 & 84.0 & - & - & ~ & - & - & - \\
			Bissaco et al.\cite{12} & - & - & - & - & ~ & 90.4 & 78.0 & - & ~ & - & - & - & - & ~ & 87.6 & - & - \\
			Goel et al.\cite{22} & - & - & - & - & ~ & 77.3 & - & - & ~ & 89.7 & - & - & - & ~ & - & - & - \\
			Alsharif \& Pineau\cite{16} & - & - & - & - & ~ & 74.3 & - & - & ~ & 93.1 & 88.6 & - & - & ~ & - & - & - \\
			Almazan et al.\cite{25} & 91.2 & 82.1 & - & - & ~ & 89.2 & - & - & ~ & - & - & - & - & ~ & - & - & - \\
			Yao et al.\cite{5} & 80.2 & 69.3 & - & - & ~ & 75.9 & - & - & ~ & 88.5 & 80.3 & - & - & ~ & - & - & - \\
			R.-Serrano et al.\cite{23} & 76.1 & 57.4 & - & - & ~ & 70.0 & - & - & ~ & - & - & - & - & ~ & - & - & - \\
			Jaderberg et al.\cite{13} & - & - & - & - & ~ & 86.1 & - & - & ~ & 96.2 & 91.5 & - & - & ~ & - & - & - \\
			Su \& Lu et al.\cite{17} & - & - & - & - & ~ & 83.0 & - & - & ~ & 92.0 & 82.0 & - & - & ~ & - & - & - \\
			Gordo\cite{24} & 93.3 & 86.6 & - & - & ~ & 91.8 & - & - & ~ & - & - & - & - & ~ & - & - & - \\
			Jaderberg et al.\cite{28} & 97.1 & 92.7 & - & - & ~ & 95.4 & - & 80.7* & ~ & \textbf{98.7} & \textbf{98.6} & - & \textbf{93.1*} & ~ & - & \textbf{90.8*} & 490M \\
			Jaderberg et al.\cite{27} & 95.5 & 89.6 & - & - & ~ & 93.2 & - & 71.7 & ~ & 97.8 & 97.0 & - & 89.6 & ~ & - & 81.8 & 304M \\
			Shi et al. \cite{20} & 97.8 & 95.0 & - & 81.2 & ~ & \textbf{97.5} & - & 82.7 & ~ & \textbf{98.7} & 98.0 & - & 91.9 & ~ & - & 89.6 & 8.3M \\
			Shi et al.\cite{19} & 96.2 & 93.8 & - & 81.9 & ~ & 95.5 & - & 81.9 & ~ & 98.3 & 96.2 & - & 90.1 & ~ & - & 88.6 & - \\
			Lee et al.\cite{29} & 96.8 & 94.4 & - & 78.4 & ~ & 96.3 & - & 80.7 & ~ & 97.9 & 97.0 & - & 88.7 & ~ & - & 90 & - \\
			\midrule\addlinespace
			Ours(n=1) & 98.6 & 96.3 & 83.0 & 80.9 & ~ & 94.4 & 82.1 & 76.7 & ~ & 97.2 & 96.0 & 88.8 & 84.1 & ~ & 88.2 & 84.9 & 8.1M \\
			Ours(n=3) & \textbf{98.9} & \textbf{96.7} & \textbf{83.5} & 81.6 & ~ & 95.1 & \textbf{84.1} & 76.5 & ~ & 97.7 & 96.4 & 90.5 & 84.5 & ~ & 89.0 & 85.2 & - \\
			Ours(n=1, Residual) & 98.7 & 96.1 & 80.6 & 78.2 & ~ & 95.1 & 79.9 & 72.5 & ~ & 97.6 & 96.5 & 87.1 & 81.1 & ~ & 86.9 & 81.4 & \textbf{0.41M} \\ \bottomrule
		\end{tabular}
	}
\end{table*}

\section{Experiments}
We implemented the model on the platform of Torch 7 \cite{36} with the CTC transcription layer (in C++) and the decoding schemes (in C++). Experiments were performed on a workstation with the Intel(R) Xeon(R) E5-2680 CPU, 256GB RAM and an NVIDIA GeForce GTX TITAN X GPU. Networks were trained with stochastic gradient descent algorithm, with the initial learning rate 0.1, and we selected $1/20$ of the training samples for each epoch. The learning rate is decreased by $\times0.3$ at the 40th epoch and the 60th epoch. The training finished in about 70 epochs.

To evaluate the effectiveness of the proposed method, we conducted experiments for English and Chinese scene texts, which are both challenging vision tasks.

\subsection{Results on English Datasets}

\subsubsection{Datasets}

For English scene text recognition, we use the synthetic dataset (Synth) released by \cite{26} as training data for all the following experiments. The training set consists of 8 millions images and their corresponding ground truth on text line level, which is generated by a synthetic data engine using a 90K word dictionary. Although the model is trained with the synthetic data only, even without any fine-tuning on specific training sets, it works well on real image datasets. We evaluated our scene text recognition system on four popular English benchmarks, namely ICDAR 2003 (IC03), ICDAR 2013 (IC13), IIIT 5k-word (IIIT5k) and Street View Text (SVT). The \textbf{IIIT5k} dataset \cite{9} contains 3,000 cropped word test images from the scene images from the Internet. Each word image has been associated to a 50-words lexicon and a 1k-words lexicon. This is the largest dataset for English scene text recognition so far. The \textbf{SVT} dataset \cite{6} was collected from Google Street View of road-side scenes. The test dataset contains 249 images, from which 647 word images were cropped, and each word image has a 50-words lexicon defined by Wang et al. \cite{6}. The \textbf{IC03} dataset \cite{37} test dataset contains 251 scene images with labeled text bounding boxes. We discard the images which either contain non-alphanumeric characters or have less than three characters following Wang et al. \cite{6}, and get a test set with 860 cropped images. Each test image is associated with a 50-words lexicon as defined by Wang et al. \cite{6}. A full lexicon is built by combining all the lexicons of per images. The \textbf{IC13} dataset \cite{39} test dataset inherits most of its data from IC03. It contains 1,015 cropped word images with ground truths.

\subsubsection{Implementation Details}

During training, all images are scaled to have height 32, widths are proportionally scaled. For parallel computation in training on GPU, we unify the normalized text images to width 256. So, If the proportional width is less than 256, we pad the scaled image to width 256, otherwhise, we continue to scale the image to $32\times256$ (this rarely happens because most words are not so long). 

In training, we investigate two types of model (single-scale model and multi-scale model). The single-scale model has only one input feature map with the window size of 32x32. While the multi-scale model has three input fature maps, which are firstly extracted with the window size of 32x24, 32x32 and 32x40, and then are all resized to 32x32. Both the models are shifted with step 4. Test images are scaled to have height 32, and the widths are proportionally scaled with heights. In testing, the window sliding step is 4 as in training.

For integrating linguistic context in recognition, we trained a 5-gram character language model (LM) on two text corpra. One is extracted from the transcripts of training imageset consisting about 8 million words. The other is trained on a general corpus \cite{41} consisting of 15 million sentences.

\subsubsection{Comparisons with State-of-the-art}
The English scene text word recognition results on four public datasets are listed in Table 2, with comparison to the state of the art. We also give results of our model using residual network \cite{42} instead of the structure specified in Section 3.2.

In the naive decoding case ({\bf None}), our method achieves comparable results with the best performance on the four datasets. It can be found that the results with three scales (n=3) are better than only one scale input (n=1), as the model can capture more context information with more scales. In the unconstrained case (w/o language model), best performance close to our work are reported by \cite{19, 27}. However, the result in \cite{27} is constrained to a 90k dictionary, and there is no out-of-vocabulary word. For the LSTM based method \cite{19}, an implicit language model is embedded training with a dataset of large lexicon (Synth contains the words of IC03/13 and SVT test set), and therefore can give higher performance. However, it is unfair to compare these methods with real lexicon-free and LM-free methods. Although we also used the Synth as training data, our model only learns character models in training. This means that our model is totally lexicon-free and LM-free, thus, it performs quite stably on different datasets when lexicon is free. In contrast, the LSTM based method \cite{19} has an obvious decrease of performance on the IIIT-5K.

In our method, the process of each window is independent, so we can parallelize the classification of all the windows in one time step. However, the LSTM-based method has each time step dependent on others, so, it must be updated step by step. Therefore, our method only need 0.015s to process each sample on average with naive decoding, whereas the average testing time is 0.16s/sample for method \cite{19}.

In the lexicon-based decoding case, our method achieves the best performance on IIIT5k, which has the largest number of cropped images. On datasets SVT and IC03, our model with lexicon-base decoding yields results comparable to the best of state of the art. 

The model size of several methods are listed in Table 2, which reports the parameter number of the learned model. The number of parameters of our model are less than all the previous deep leaning methods \cite{19, 27, 28}. Moreover, our model size can be largely reduced to only 0.41M by a 38-layer residual network \cite{42}, while the performance can keep comparable to our convolutional character model. This is a good trade off between the space and accuracy and can be easily ported to mobile devices.

\begin{figure*}
	\hspace{2em}
	\begin{subfigure}[h]{0.505\textwidth}
		\centering
		\includegraphics[width=0.9\textwidth,height=0.8\textwidth]{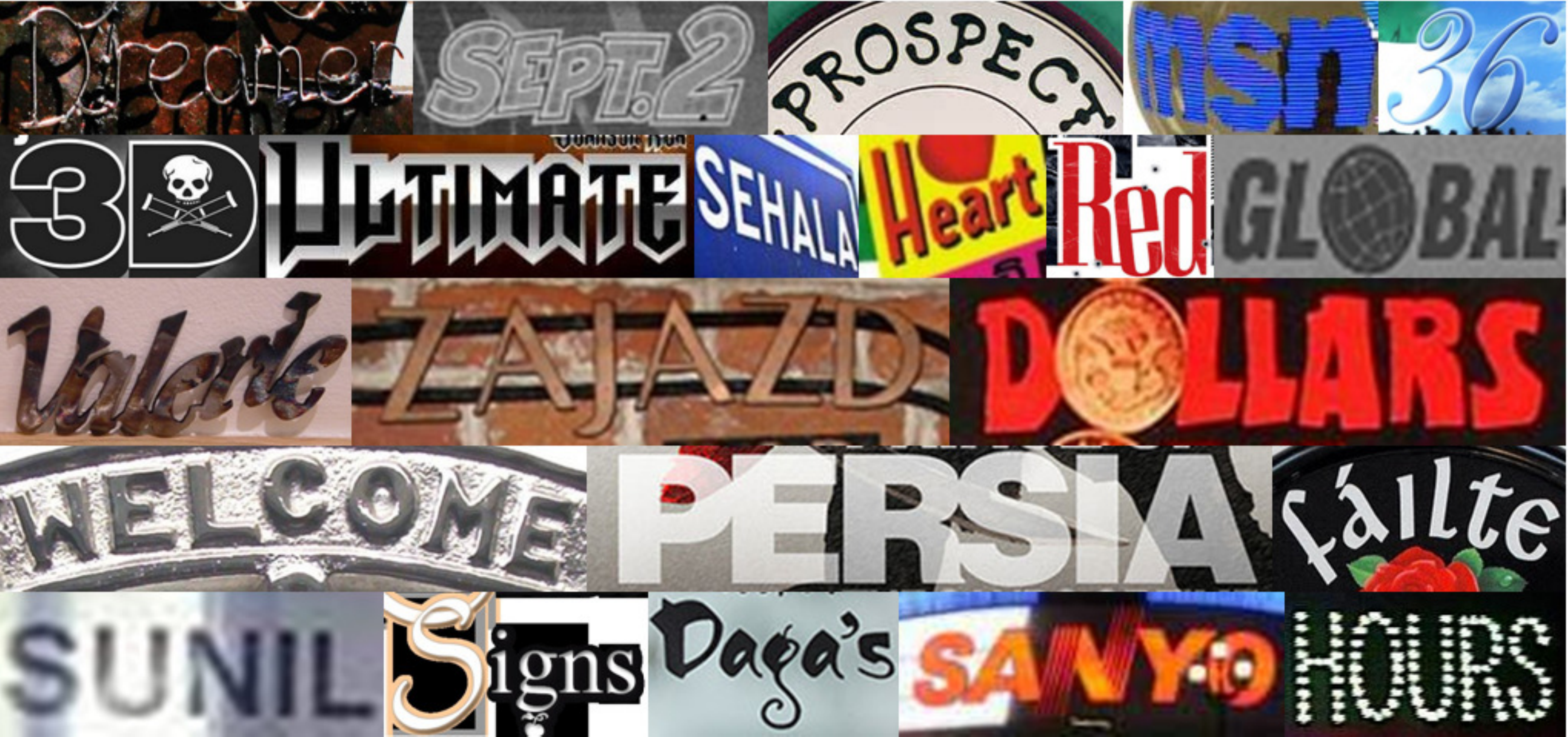}
		\subcaption{~}
	\end{subfigure}
	\hfill
	\begin{subfigure}[h]{0.475\textwidth}
		\centering
		\includegraphics[width=0.6\textwidth,height=0.85\textwidth]{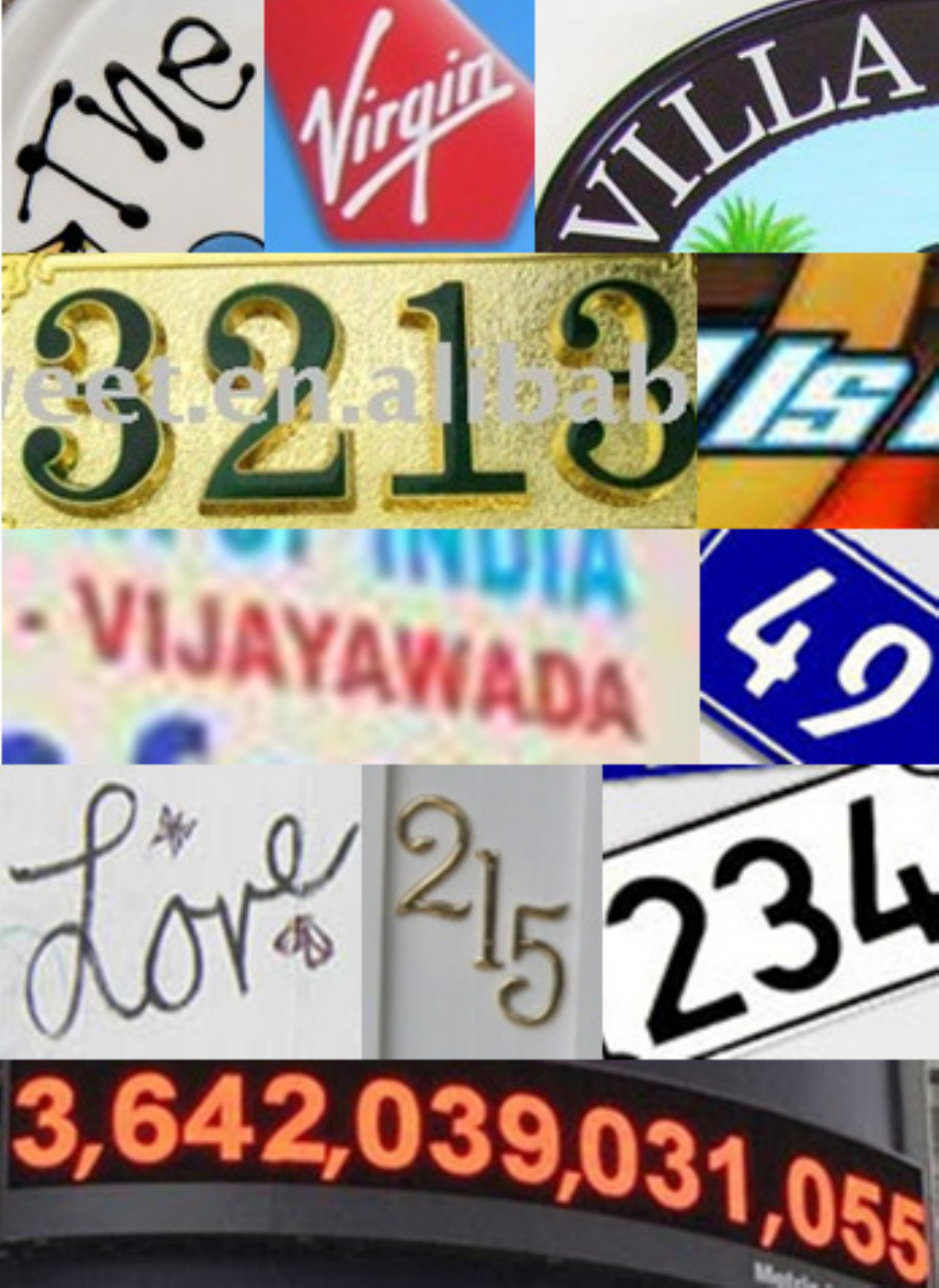}
		\subcaption{~}
	\end{subfigure}
	\caption[ ]
	{\small (a) Correct recognition samples in IIIT5k; (b) Incorrect recognition samples in IIIT5k}
	\label{fig:mean and std of nets}
	\hspace{3em}
\end{figure*}

\subsection{Results on Chinese Datasets}

\subsubsection{Datasets}
In the training set of the public available Chinese text recognition dataset \cite{45}, there are only about 1400 character classes. However, with that we cannot train a practicable Chinese scene text recognition system because the common used simplified Chinese characters are actually more than 7000 classes. Therefore, following some success syntentic text dataset, we generated a Chinese scene text dataset (\textbf{Synth-Ch}) with the engine in \cite{26}. The Synth-Ch includes 9 million text images, containing the number of characters from 1 to 15 in every text image. Moreover, for each text image, the font of characters are randomly selected from 60 Chinese fonts, and the characters are randomly selected from a Chinese character list which includes 7,185 common used simplified characters and 171 symbols (including 52 English letters and 10 digits). 

For Chinese scene text recognition, we use the Synth-Ch as training data. We evaluated our Chinese scene text recognition system on ICDAR2015 Text Reading in the Wild Competition dataset (TRW15). The TRW15 dataset \cite{45} contains 984 images and 484 images are selected as test set. From the testing images, we cropped 2996 horizontal text lines as the first test set (\textbf{TRW15-T}). However, the number of character classes is small (about 1460 character classes) in TRW15-T. Thus, in order to evaluate the performance of our model more efficinetly, we constructed the second test set (\textbf{TRW15-A}) from all 984 images. In TRW15-A, there are 6106 horizontal text lines and more than 1800 character classes.

\subsubsection{Implementation Details}

During training, we use the same normalization strategy as that in English data sets, except the width of normalized text images is set as 512. we only train single-scale model with the window size of 32x40. In training, the shifted step is set as 8. Test images are firstly rectified to a rectangle image with perspective transformation \cite{47} because most of the test images have perspective distort, then the rectified image are scaled to have height 32, and the widths are proportionally scaled with heights. In testing, the window sliding step is 8 as in training. We evaluate the recognition performance using character-level accuracy (Accurate Rate) following \cite{2}. 

For integrating linguistic context in recognition, we trained a 5-gram character language model (LM) on the SLD corpus \cite{46}, which contains news text from the 2006 Sogou Labs data.

\subsubsection{Comparisons with State-of-the-art}

The Chinese scene text recognition results on TRW15 datasets are listed in Table 3. Compared to previous results reported on the ICDAR2015 Text Reading in the Wild Competition, 72.1 percent of AR, the proposed approach achieved 81.2 percent of AR, demonstrating significant improvement and advantage. We also give results of our model using residual network \cite{42} instead of the structure specified in Section 3.2, which is also much better than the winner method in the competition.

\begin{table}[!htb]
	\centering
	\caption{Recognition accuracies (\%) on Chinese scene text dataset}
	\scalebox{0.7}[0.7]{
		\begin{tabular}{ccccccccc}\toprule
			~ & \multicolumn{2}{c}{\textbf{TRW15-T}} & ~ & \multicolumn{2}{c}{\textbf{TRW15-A}} & ~ \\
			\cmidrule(lr){2-3}
			\cmidrule(lr){5-6}
			
			\textbf{Method} & \textbf{LM} & \textbf{None} & ~ & \textbf{LM} & \textbf{None} & ~ & \textbf{Model Size} \\ \midrule
			baseline \cite{45} & 26.5 & - & ~ & - & - & ~ &  - \\
			CASIA-NLPR \cite{45} & 72.1 & - & ~ & - & - & ~ & 148M \\
			\midrule\addlinespace
			Ours(n=1) & \textbf{81.2} & 76.5 & ~ & \textbf{81.7} & 76.8 & ~ & 9.6M  \\
			Ours(n=1, Residual) & 77.9 & 71.3 & ~ & 78.2 & 71.6 & ~ & \textbf{2.3M} \\ \bottomrule
		\end{tabular}
	}
\end{table}

\section{Conclusion}
In this paper, we investigate the intrinsic characteristics of text recognition, and inspired by human cognition mechanisms in reading texts, we propose a scene text recognition method with character models on convolutional feature map. The model is trained end-to-end on word images weakly labeled with transcripts. The experiments on English and Chinese scene text recognition demonstrate that the proposed method achieves superior or comparable performance. In the future, we will evaluate our model on more challenging data sets. 



{\small
	\bibliographystyle{ieee}
	\bibliography{egbib}
}

\end{document}